\documentclass[10pt,twocolumn,letterpaper]{article}

\usepackage{cvpr}
\usepackage{url}
\usepackage{lineno}
\definecolor{White}{rgb}{1.,0.,1.}
\definecolor{first}{rgb}{.8,.0,.0}
\definecolor{second}{rgb}{.0,.6,.0}
\definecolor{third}{rgb}{.0,.0,.8}

\definecolor{ceiling}{RGB}{214,  38, 40}
\definecolor{floor}{RGB}{43, 160, 4}
\definecolor{wall}{RGB}{158, 216, 229}
\definecolor{window}{RGB}{114, 158, 206}
\definecolor{chair}{RGB}{204, 204, 91}
\definecolor{bed}{RGB}{255, 186, 119}
\definecolor{sofa}{RGB}{147, 102, 188}
\definecolor{table}{RGB}{30, 119, 181}
\definecolor{tvs}{RGB}{160, 188, 33}
\definecolor{furniture}{RGB}{255, 127, 12}
\definecolor{objects}{RGB}{196, 175, 214}

\definecolor{car}{rgb}{0.39215686, 0.58823529, 0.96078431}
\definecolor{bicycle}{rgb}{0.39215686, 0.90196078, 0.96078431}
\definecolor{motorcycle}{rgb}{0.11764706, 0.23529412, 0.58823529}
\definecolor{truck}{rgb}{0.31372549, 0.11764706, 0.70588235}
\definecolor{othervehicle}{rgb}{0.39215686, 0.31372549, 0.98039216}
\definecolor{person}{rgb}{1.        , 0.11764706, 0.11764706}
\definecolor{bicyclist}{rgb}{1.        , 0.15686275, 0.78431373}
\definecolor{motorcyclist}{rgb}{0.58823529, 0.11764706, 0.35294118}
\definecolor{road}{rgb}{1.        , 0.        , 1.        }
\definecolor{parking}{rgb}{1.        , 0.58823529, 1.        }
\definecolor{sidewalk}{rgb}{0.29411765, 0.        , 0.29411765}
\definecolor{otherground}{rgb}{0.68627451, 0.        , 0.29411765}
\definecolor{building}{rgb}{1.        , 0.78431373, 0.        }
\definecolor{fence}{rgb}{1.        , 0.47058824, 0.19607843}
\definecolor{vegetation}{rgb}{0.        , 0.68627451, 0.        }
\definecolor{trunk}{rgb}{0.52941176, 0.23529412, 0.        }
\definecolor{terrain}{rgb}{0.58823529, 0.94117647, 0.31372549}
\definecolor{pole}{rgb}{1.        , 0.94117647, 0.58823529}
\definecolor{trafficsign}{rgb}{1.        , 0.        , 0.        }
\definecolor{otherstructure}{rgb}{0.98039215, 0.58823529, 0.}
\definecolor{otherobject}{rgb}{0.19607843, 1.        , 1.        }

\makeatletter
\newcommand{\car@semkitfreq}{3.92}
\newcommand{\bicycle@semkitfreq}{0.03}
\newcommand{\motorcycle@semkitfreq}{0.03}
\newcommand{\truck@semkitfreq}{0.16}
\newcommand{\othervehicle@semkitfreq}{0.20}
\newcommand{\person@semkitfreq}{0.07}
\newcommand{\bicyclist@semkitfreq}{0.07}
\newcommand{\motorcyclist@semkitfreq}{0.05}
\newcommand{\road@semkitfreq}{15.30}
\newcommand{\parking@semkitfreq}{1.12}
\newcommand{\sidewalk@semkitfreq}{11.13}
\newcommand{\otherground@semkitfreq}{0.56}
\newcommand{\building@semkitfreq}{14.1}
\newcommand{\fence@semkitfreq}{3.90}
\newcommand{\vegetation@semkitfreq}{39.3}
\newcommand{\trunk@semkitfreq}{0.51}
\newcommand{\terrain@semkitfreq}{9.17}
\newcommand{\pole@semkitfreq}{0.29}
\newcommand{\trafficsign@semkitfreq}{0.08}
\newcommand{\semkitfreq}[1]{{\csname #1@semkitfreq\endcsname}}

\newcommand{\car@sscbkitfreq}{2.85}
\newcommand{\bicycle@sscbkitfreq}{0.01}
\newcommand{\motorcycle@sscbkitfreq}{0.01}
\newcommand{\truck@sscbkitfreq}{0.16}
\newcommand{\othervehicle@sscbkitfreq}{5.75}
\newcommand{\person@sscbkitfreq}{0.02}
\newcommand{\road@sscbkitfreq}{14.98}
\newcommand{\parking@sscbkitfreq}{2.31}
\newcommand{\sidewalk@sscbkitfreq}{6.43}
\newcommand{\otherground@sscbkitfreq}{2.05}
\newcommand{\building@sscbkitfreq}{15.67}
\newcommand{\fence@sscbkitfreq}{0.96}
\newcommand{\vegetation@sscbkitfreq}{41.99}
\newcommand{\terrain@sscbkitfreq}{7.10}
\newcommand{\pole@sscbkitfreq}{0.22}
\newcommand{\trafficsign@sscbkitfreq}{0.06}
\newcommand{\otherstructure@sscbkitfreq}{4.33}
\newcommand{\otherobject@sscbkitfreq}{0.28}
\newcommand{\sscbkitfreq}[1]{{\csname #1@sscbkitfreq\endcsname}}

\definecolor{cvprblue}{rgb}{0.21,0.49,0.74}
\usepackage[pagebackref,breaklinks,colorlinks,citecolor=cvprblue]{hyperref}

\def\paperID{8934} 
\def\confName{CVPR}
\def\confYear{2025}

\title{SGFormer: Satellite-Ground Fusion for 3D Semantic Scene Completion}

\author{Xiyue Guo$^{1}$ $^{\ast}$ \quad Jiarui Hu$^{1}$ $^{\ast}$ \quad Junjie Hu$^{2}$ \quad Hujun Bao$^{1}$ \quad Guofeng Zhang$^{1\dagger}$\\
$^{1}$State Key Lab of CAD\&CG, Zhejiang University \quad  $^{2}$Chinese University of Hong Kong, Shenzhen
}
\begin{document}

\twocolumn[{%
\renewcommand\twocolumn[1][]{#1}%
\maketitle
\begin{center}
    \centering
    \vspace{-6mm}
    \captionsetup{type=figure}
    \includegraphics[width=\linewidth]{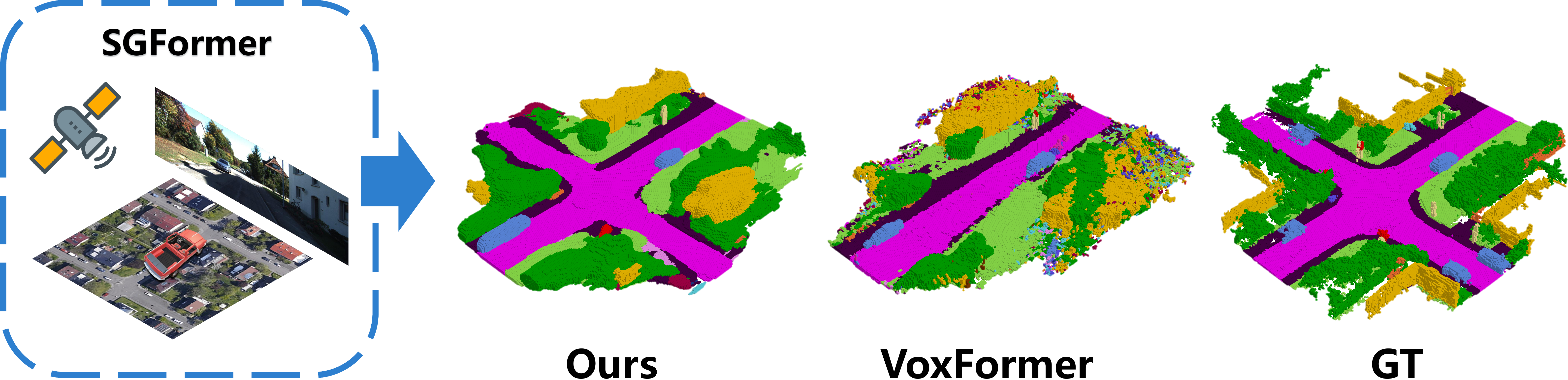}
    \vspace{-6mm}
    \captionof{figure}{\textbf{SGFormer}, which adopts satellite-ground cooperative fusion, can achieve state-of-the-art performance in scene completion and semantic prediction. Benefiting from informative satellite images and a well-designed dual-branch pipeline, SGFormer can effectively improve semantic prediction accuracy and solve the long-standing visual occlusion bottleneck suffered by purely ground-view methods.}
    \label{fig:teaser}
    \vspace{-3mm}
\end{center}%
}]
\renewcommand{\thefootnote}{\fnsymbol{footnote}} 
\footnotetext{$^{\dagger}$Corresponding author. Email: zhangguofeng@zju.edu.cn}
\footnotetext{$^{\ast}$ Equal Contribution.}

\begin{abstract}
Recently, camera-based solutions have been extensively explored for scene semantic completion (SSC). Despite their success in visible areas, existing methods struggle to capture complete scene semantics due to frequent visual occlusions. 
To address this limitation, this paper presents the first satellite-ground cooperative SSC framework, \,i.e., SGFormer, exploring the potential of satellite-ground image pairs in the SSC task. 
Specifically, we propose a dual-branch architecture that encodes orthogonal satellite and ground views in parallel, unifying them into a common domain. Additionally, we design a ground-view guidance strategy that corrects satellite image biases during feature encoding, addressing misalignment between satellite and ground views. Moreover, we develop an adaptive weighting strategy that balances contributions from satellite and ground views. Experiments demonstrate that SGFormer outperforms the state of the art on SemanticKITTI and SSCBench-KITTI-360 datasets. Our code is available on \url{https://github.com/gxytcrc/SGFormer}.

\end{abstract}
    
\section{Introduction}
\label{sec:intro}
Urban scene semantic completion (SSC) has been an increasingly prominent problem in 3D computer vision over recent decades, which targets predicting 3D semantic and geometric occupancy of immediately observed surroundings with a variety of downstream applications such as autonomous driving, robot navigation, and augmented reality (AR). Lidar-based methods have achieved remarkable progress and decent performance~\cite{sscnet, js3cnet, lmscnet}, while the underlying point cloud representation inherently suffers from weak semantic context derived only from geometry shapes.  In contrast, cost-effective camera-based methods can lift rich 2D cues into the 3D world, demonstrating potential in scene reconstruction and understanding. 

Existing camera-based methods have shown promising performance on the SSC task \cite{tpvformer, surroundocc,voxformer,hayler2024s4c, CGFormer}. Nevertheless, even with depth information for visible areas, they still face the challenge of non-unique correspondences between 3D volumes and 2D pixels. Specifically, multiple 3D volumes correspond to significantly overlapping regions within the 2D image plane, which causes semantic ambiguity and radial artifacts in the final reconstruction. For occluded areas, these methods generally lack a long-range global perspective, which makes them struggle to restore a complete scene and serve the following planning and decision steps.

In this paper, for the first time, we propose to incorporate satellite imagery into the SSC task.
Alongside the development of remote sensing technology, satellite imagery has emerged as a low-cost and widely available reference information. A satellite image, covering major traffic flows in a city, typically requires a light-weight storage footprint, which makes it a compact and memory-efficient representation. Meanwhile, the bird's-eye view (BEV) observation is highly suitable for the horizontal layout of urban scenes. It provides a broad perspective of surrounding obvious objects to effectively enhance semantic certainty. These two orthogonal views, the satellite and ground view, can optimally compensate for each other’s blind fields and provide an ideal solution to the previous occlusion bottleneck.

However, there are two difficulties in incorporating satellite imagery into the SSC task. The first is the misalignment issue. A local satellite image is captured as a fixed-size 2D segment centered around a specific location. An unpredictable deviation from noisy localization and top-down occlusions in the satellite image are the root causes of this issue. Misaligned satellite features tend to bring feature-level inconsistency in fusion and have a negative impact on convergence efficiency. Second, satellite images are typically pre-captured and mainly focus on the scene layout, while inevitably discarding details and long-term changes, such as traffic signs and temporarily parked vehicles, and degrading SSC performance in such low-occupancy but important regions. 

To solve the above challenges, we propose SGFormer, a tightly coupled satellite-assisted framework for the SSC task.
In order to perform satellite-ground cross-view fusion, SGFormer innovatively proposes a tailored dual-branch framework to synchronously encode ground and satellite images, aligning them within a unified feature domain.
To overcome the misalignment issue, we design a feature-level satellite correction strategy regarding vertically squeezed ground-view features. Specifically, we initialize the learnable BEV parameters and iteratively query squeezed ground-view features in the deformable self-attention layer, where ground-view guidance plays a crucial role to warm up BEV queries in advance to enable coordinated fusion. 
In addition, we propose an adaptive fusion module with a dual-path weight generator to achieve a reasonable trade-off between two orthogonal views. Within this module, the concatenated satellite-ground features are decoded into channel-wise weight vectors for both views in each voxel. This module allows our fusion pipeline to handle temporal updates and objects of various sizes.

We evaluate our method on two benchmark semantic datasets, including SemanticKITTI~\cite{semantickitti} and SSCBench-KITTI-360~\cite{sscbench}. Extensive experimental results demonstrate that our SGFormer outperforms previous approaches, yielding superior performance in terms of scene completion and semantic prediction, as shown in Figure~\ref{fig:teaser}. Overall, our contributions can be summarized as follows:

\begin{itemize}[leftmargin=2em]
    \item We present the first tightly coupled satellite-assisted SSC framework with a dual-branch design tailored for satellite observations.
    \item We propose a feature-level satellite correction strategy based on deformable self-attention to address the misalignment between satellite and ground views.
    \item We develop an adaptive weighting strategy to balance contributions from satellite and ground view, enabling effective perception of dynamic updates and objects of different sizes.
    \item  Experiments on various datasets illustrate that our method can achieve better scene semantic completion results compared to baselines.
\end{itemize}
\begin{figure*}
  \centering
  \includegraphics[width=\linewidth]{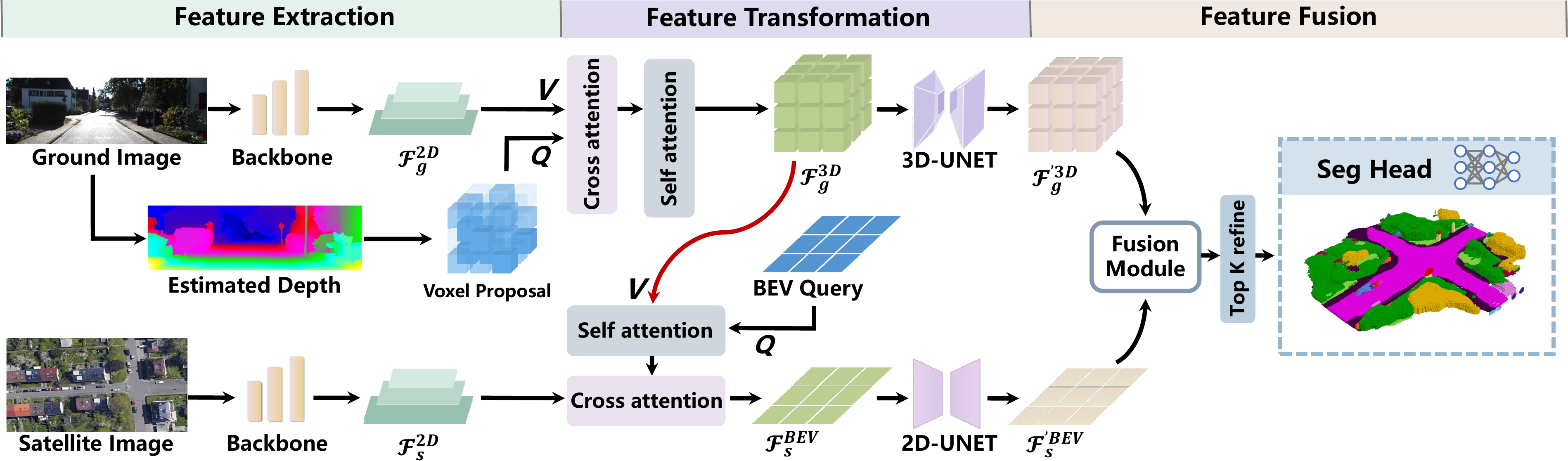}
  \vspace{-5mm}
  \caption{\textbf{Overview of SGFormer.} Overall, SGFormer feeds a satellite-ground image pair into similar backbone networks in different branches to extract multi-level feature maps respectively \textbf{(Left Part)}. Then, leveraging deformable attention, it transforms satellite and ground features into volume and BEV spaces \textbf{(Middle Part)} for following feature fusion and decoding \textbf{(Right Part)}. Specifically, in the ground branch, we use a depth estimator to produce voxel proposals for targeted querying on non-empty feature volumes. In the satellite branch, we fuse vertically squeezed ground-view features into BEV queries to warm up satellite features (\textcolor[rgb]{0.7,0,0}{Red Line}). Before final fusion, encoded features from both branches are enhanced through 2D/3D convolution networks. Our proposed fusion module, which is detailed in Figure~\ref{fig:fusion}, is able to adaptively fuse satellite and ground features, followed by a seg head to output semantic reconstruction results.}
  \label{fig:framework}
  \vspace{-4mm}
\end{figure*}

\section{Related Work}
\subsection{3D Semantic Scene Completion}
3D Semantic Scene Completion (SSC) was first proposed by SSCNet~\cite{sscnet}, aiming to predict the occupancy and semantic information of each voxel in 3D spaces~\cite{s3cnet,js3cnet,lmscnet,scpnet}. Subsequent methods can be categorized into two types: one relies on depth sensors, such as LiDAR, to directly compute spatial features for estimating 3D semantics, while the other is camera-based and involves lifting image features to the 3D spaces before estimating semantic information. Earlier approaches mainly focused on the first type, while in recent years, camera-based solutions have gained increasing attention due to their high cost-efficiency~\cite{monoscene, tpvformer, surroundocc, voxformer}.

MonoScene \cite{monoscene} presents the first pure vision solution, leveraging 2D and 3D U-Nets bridged by a sight projection module. OccFormer \cite{occformer} follows the strategy of LSS \cite{lss}, estimates the depth distribution of the image, and then projects the features into 3D spaces through depth guidance.

Some works are inspired by BEVFormer \cite{bevformer}, which utilizes the Transformer framework to estimate occupancy. These methods use spatial relationships from 3D to 2D to query the information from image features via spatial deformable attention~\cite{deformabledetr}. Among them, TPVFormer~\cite{tpvformer} proposes a Tri-perspective view representation. It first acquires the features of these planes and recovers them into 3D accordingly. SurrunodOcc~\cite{surroundocc} introduces a coarse-to-fine strategy, generating 3D features at multiple scales and progressively integrating occupancy grid predictions. These transformer-based methods show significant progress in the performance of 3D semantic prediction. However, due to the non-one-to-one nature of the 3D-2D projection relationship and the lack of geometric constraints, these methods struggle to reconstruct the semantic distribution of 3D scenes accurately. 

To address the problem, some works leverage depth input and try to incorporate geometric priors into occupancy predictions. OccDepth~\cite{occdepth} projects the image features by stereo depth. Voxformer~\cite{voxformer} initializes the sparse proposal from pre-trained depth. Subsequent studies~\cite{symphonize, sparseocc, depthcite} have further improved upon this by integrating in-depth information, thereby enhancing the performance of SSC. However, due to inherent issues arising from 
limitations of ground camera observation range and occlusions make predictive semantic ambiguities and geometric distribution errors almost inevitable.

\subsection{Satellite Assist Perception}
Recently, integrating satellite images with ground images has gained increasing attention. This is mainly due to the low economic and storage costs of satellite images, as well as the wealth of information they contain.

Most of these works mainly focus on the localization problem, aiming at estimating the position of ground vehicles in real-world coordinates by matching ground and satellite images. Some methods divide satellite maps into many small patches, aiming to find the patch that is most similar to the ground image through image retrieval methods \cite{satretrive1, satretrive2, satretrive3, satretrive4}. Other approaches try integrating satellite and ground features into the common coordinate system to achieve higher localization accuracy, such as applying homography to project satellite features into the ground perspective and then determining precise localization results \cite{beyond} or projecting ground images into a bird's-eye view (BEV) and aligning them with satellite features~\cite{ori, uncertainty, guo}. On this basis, some works have attempted to incorporate the results of such cross-view localization into traditional SLAM systems to enhance SLAM localization performance~\cite{guo, anyway, increasing}.

In addition to localization efforts, some works have incorporated satellite images into map construction tasks. SG-BEV~\cite{sg-bev} is the first to combine ground and satellite features to accomplish fine-grained building attribute segmentation tasks, while SNAP~\cite{sarlin2024snap} included satellite images in the construction of 2D neural maps.

Against this backdrop, our approach is the first to introduce satellite images into the SSC task, exploring the potential of cooperative satellite and ground perception.


\section{Method}
\subsection{Overview}
The framework of our proposed SGFormer is illustrated in Figure~\ref{fig:framework}. Our work takes both ground and satellite images as inputs, processed through two distinct branches: the ground branch and the satellite branch. Each branch consists of 
two-step operations, where the first is the feature extraction to generate muti-scale features from ground and satellite images and the second is the feature transformation to covert the features into voxel or BEV spaces, followed by feature diffusion and enhancement. Then, the voxel and BEV features are fused through a fusion module. 

Within SGFormer, we  make following main technical contributions:
1) Satellite-view feature correction: the satellite correction method employs compressed features from the ground branch to guide and correct the feature learning of the satellite branch (detailed in Sec.~\ref{subsec_satellite_view_fea}).
2) Adaptive fusion: The adaptive fusion module merges BEV features from the satellite branch with voxel features from the ground branch through the attention mechanism. After fusing features from both branches, we further refine them and pass the output into a segmentation head, which upsamples these features and produces voxel-wise class predictions (detailed in Sec.~\ref{subsec_feature_fusion}).

\subsection{Feature Extraction.} Feature extraction of the ground branch aims to extract 2D features $\mathbf{F}_g^{2D} \in \mathbb{R} ^{H_g' \times W_g' \times D}$ from the ground image, where $H_g' \times W_g'$ represents the feature resolution, and $D$ is the feature dimension. Similarly, in the satellite branch, we obtain 2D satellite features \textbf{$\mathbf{F}_s^{2D} \in \mathbb{R} ^{H_s' \times W_s' \times D}$}. EfficientNet-B7~\cite{efficientnet} serves as the backbone for ground images, while ResNet-50~\cite{resnet} is used for satellite images, with both backbones followed by a Feature Pyramid Network (FPN)~\cite{FPN}. 

\noindent \textbf{Depth Estimation.} Before encoding the features of ground view to 3D spaces, we estimate the visible voxels using depth estimation, employing the pre-trained MobileStereoNet~\cite{mobilestereonet} for depth estimation, aligning to VoxFormer~\cite{voxformer}. However, instead of using an additional stage to refine the binary occupancy like VoxFormer~\cite{voxformer}, we directly employ the voxelized depth as our query proposals for feature encoding. Specifically, depth points are projected into the voxel map, with a voxel set to 1 if the number of occupied points exceeds a threshold; otherwise, it is set to 0.

\begin{figure}[t]
  \centering
   \includegraphics[width=\linewidth]{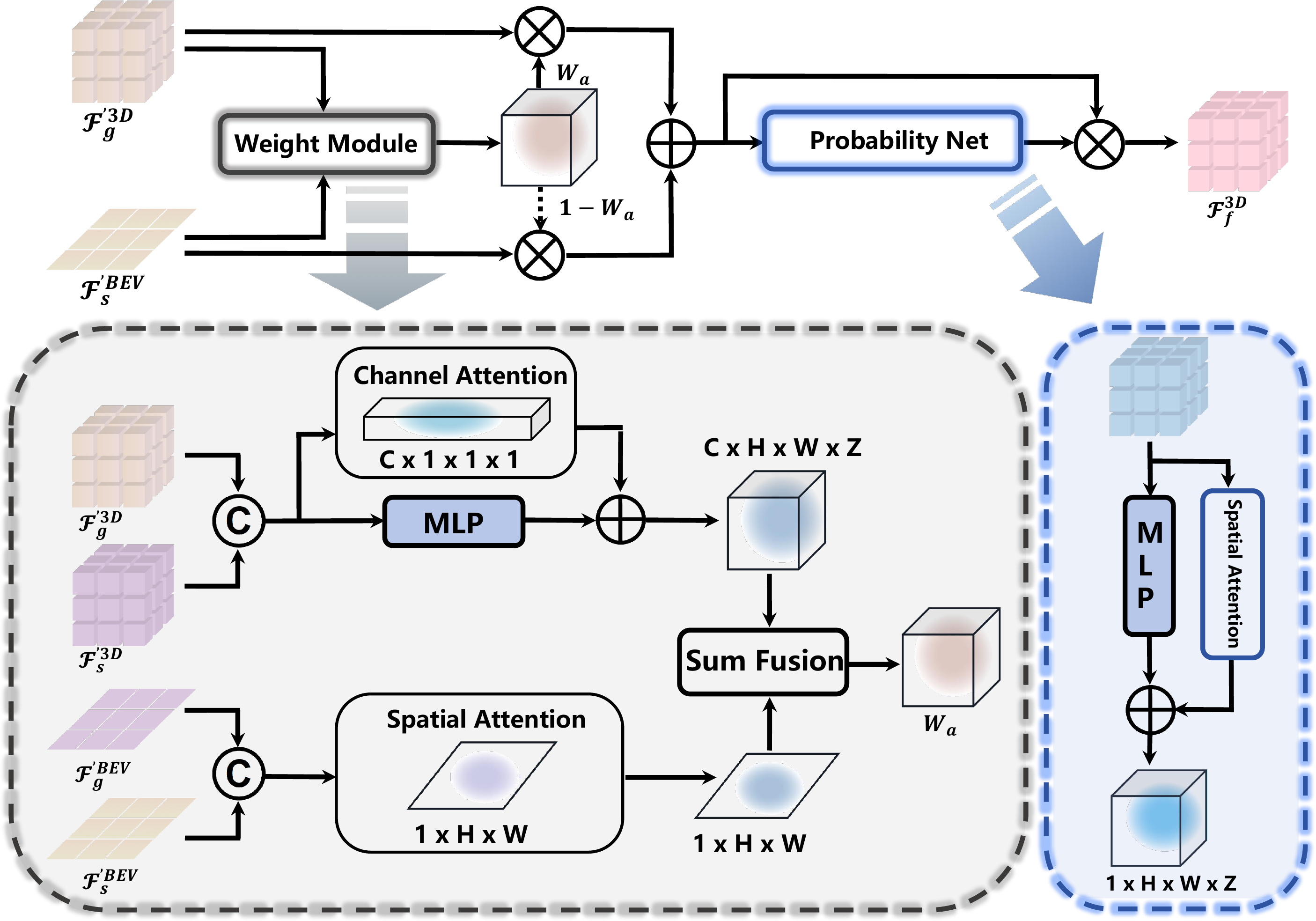}
   \vspace{-4mm}
   \caption{\textbf{Fusion Module}. Our fusion module \textbf{(Top)} mainly consists of two parts: the weight module and the probability network. This weight module \textbf{(Bottom Left)} can facilitate mutual perception between the BEV feature $\mathbf{F'}_s^{BEV}$ and voxel feature $\mathbf{F'}_g^{3D}$ in both 2D and 3D spaces to produce channel-wise weight vectors. This probability network \textbf{(Bottom Right)} takes the weighted average 3D voxel feature as input to infer the approximate occupancy probability per voxel to achieve better geometry reconstruction.}
   \label{fig:fusion}
   \vspace{-4mm}
\end{figure}

\subsection{Feature Transformation}
To efficiently transform the image features to real-world coordinates, we use deformable attention~\cite{deformabledetr}, represented as $\text{DA}$, to aggregate query features $q$ from target image features $v$, as described by the equation:
\begin{equation}
\operatorname{DA}\left(q, p, v\right)=\sum_{k=1}^K A_{k} W v\left(p+\Delta p\right),
\end{equation}
where p is the reference point, $\Delta p$ is the learnable sampling offset, $A_k$ represents the attention weight, and $W$ denotes the projection weight.  

\subsubsection{Ground-view Feature Transformation}
In the ground branch, we initialize the voxel queries $\mathbf{Q}_v \in \mathbb{R} ^{H \times W \times Z \times D}$ with learnable embeddings, where $H \times W \times Z$ is regarded as the voxel dimension. Guided by the query proposals, we query the voxel features from image features $\mathbf{F}_g^{2D}$ by employing the deformable cross-attention:
\begin{equation}
\mathbf{Q}_p=\operatorname{DA}\left(\mathbf{Q}_p, \mathcal{P}(\mathbf{p}, g), \mathbf{F}_g^{2 D}\right),
\end{equation}
where the $\mathbf{Q}_p$ consists of the visible voxels selected by the query proposals. $\mathcal{P}(\mathbf{p}, g)$ denotes the corresponding pixels in $F_g^{2D}$ generated by camera projection function.

After several layers of cross-attention, we merge the query features $Q_p$ with other invisible voxels in $Q_v$ to get the voxel features $\mathbf{F}_g^{3D}$, and then diffuse the features to all voxels through deformable self-attention, 
as formulated as:
\begin{equation}
\mathbf{F}_g^{3D}=\operatorname{DA}\left(\mathbf{F}_g^{3D}, \mathbf{p_v}, \mathbf{F}_g^{3D}\right),
\end{equation}
where $\mathbf{p}_v$ is the reference point in 3D space.

\subsubsection{Satellite-view Feature Transformation}
\label{subsec_satellite_view_fea}
In the satellite branch, we encode the satellite features to BEV space under the guidance of ground observations. We first initialize the BEV queries  $\mathbf{Q}_{bev} \in \mathbb{R} ^{H \times W \times D}$. Simultaneously, we compress the voxel features $\mathbf{F}_g^{3D}$ from the ground branch into BEV form $\mathbf{F}_g^{BEV}$ through max pooling. We then combine $\mathbf{Q}_{bev}$ and $\mathbf{F}_g^{BEV}$ to form a hybrid feature $v_{hybrid}$, which is fed into a self-attention module:
\begin{equation}
\mathbf{Q}_{bev}=\operatorname{DA}\left(\mathbf{Q}_{bev}, \mathbf{p}_{bev}, v_{hybrid}\right),
\end{equation}
where $\mathbf{\mathbf{p}_{bev}}$ is the reference point in BEV space. Through self-attention, the BEV queries are efficiently fused from ground-view information. The output queries are then passed into the cross-attention module to query the information from the satellite features $\mathbf{F}_s^{2D}$:
\begin{equation}
\mathbf{Q}_{bev}=\operatorname{DA}\left(\mathbf{Q}_{bev}, \mathcal{P}(\mathbf{p}, s), \mathbf{F}_s^{2D}\right),
\end{equation}
$\mathcal{P}(\mathbf{p}, s)$ represents the corresponding pixels of BEV grids in the satellite image. With the ground-view observation, the offset layers predict more suitable offsets during cross-attention. We iterate the self and cross attention several times, resulting in the BEV features $\mathbf{F}_s^{BEV}$.

\subsubsection{Convolutional Enhancement}
After encoding the features into voxel and BEV spaces, we feed them into convolutional layers to further enhance feature representations through neighborhood interactions. We use 3D-UNet for feature aggregation in the ground branch and 2D-UNet in the satellite branch.

\subsection{Feature Fusion}
\label{subsec_feature_fusion}
Our adaptive fusion module is shown in Figure~\ref{fig:fusion}. The BEV features $\mathbf{F'}_s^{BEV}$ of the satellite branch can cover a wider range, allowing for a more complete representation of scene layouts such as roads and buildings. However, BEV features lack occupancy geometry (empty or occupied) and do not perform well on small objects. On the other hand, voxel features $\mathbf{F'}_g^{3D}$ based on ground images have significant advantages in handling small, dynamic objects and detailed elements, as well as spatial occupancy, but the effective information they contain is limited to the unoccluded field of view. Therefore, it is important to keep the advantages of each view while minimizing the impact of their limitations. 

To address the challenges, we propose a fusion module that dynamically predicts weights for both the feature channel and spatial domains. Specifically, we process the features from two branches: a channel-attention network~\cite{SE} and a spatial-attention network to balance the contributions at the object level and perception filed level.

In the channel-attention path, we first lift the BEV features $\mathbf{F'}_s^{BEV}$ to 3D volumes, denoted as $\mathbf{F'}_s^{3D}$. Then, we concatenate the features from the ground and satellite branches in 3D volume to form the combined features and pass them into the channel-attention network, resulting in channel-domain weights $\mathbf{W}_c \in \mathbb{R}^{D \times 1 \times 1 \times 1}$.

In the spatial-attention path, we compress the voxel features $\mathbf{F'}_g^{3D}$ along the z-axis into BEV space, resulting in $\mathbf{F'}_g^{BEV}$. Similarly to the channel-attention path, we concatenate the BEV features and pass them into the spatial-attention network, generating spatial-domain weights $\mathbf{W}_s \in \mathbb{R}^{1 \times H \times W}$ in BEV space.
The two attention weights are summed together and then combined with the result from an additional MLP to obtain the fused weight $\mathbf{W}_a \in \mathbb{R}^{D \times H \times W \times Z}$, expressed as follows:
\begin{equation}
    \small
    \mathbf{W}_a = \operatorname{MLP}(\mathbf{F'}_c^{3D})\oplus\operatorname{C}(\mathbf{F'}_c^{3D}) \oplus \operatorname{S}(\mathbf{F'}_c^{BEV}),
\end{equation}
where $\operatorname{C}$ and $\operatorname{S}$ represent the channel and spatial attention networks, respectively. $\mathbf{F'}_c^{3D}$ and $\mathbf{F'}_c^{BEV}$ are the concatenated features in 3D and BEV spaces. $\operatorname{MLP}$ refers to the MLP operation applied to $\mathbf{F'}_c^{3D}$.
After obtaining the attention weight, we fuse the two branches' features by equation:
\begin{equation}
    \mathbf{F}_f^{3D} = \mathbf{W}_a \cdot \mathbf{F'}_g^{3D} + (1- \mathbf{W}_a) \cdot \mathbf{F'}_s^{3D},
\end{equation}
where $\mathbf{F}_f^{3D}$ is the fused voxel features.

Additionally, to minimize the negative impact of empty voxels, we get inspiration from \cite{dualbev} and \cite{cbam}, applying a probability network along with spatial attention to identify valuable voxels. Those valuable voxels will get a higher weight through the network to enhance learning efficiency. 

\noindent \textbf{Feature Refinement.}
We further refine the fused features before we output the final voxel prediction. Instead of using dense operations to refine all grids, we only focus on the grids with high uncertainty. Therefore, we first project the features $\mathbf{F}_f^{3D}$ into semantic classes $\mathbf{L}_{coarse} \in \mathbb{R} ^{N\times H \times W \times Z}$, where N is the classes number. We then compute the entropy of each grid and select top-k grids with the highest entropy scores. These high-uncertainty voxels resample features from ground features $\mathbf{F}_g^{2D}$ through deformable cross-attention. The refined features are then upsampled and output through our semantic head.
\subsection{Training Loss}
\label{sec:loss}
Following \cite{monoscene} and \cite{voxformer}, we train our model with weighted cross-entropy loss $\mathcal{L}_{ce}$, as well as scene class affinity loss $\mathcal{L}_{scal}^{geo}$ and $\mathcal{L}_{scal}^{sem}$. Moreover, in order to enhance the supervision, we additionally apply the BEV loss $\mathcal{L}_{bev}$ from the satellite branch and coarse loss $\mathcal{L}_{co}$  from the uncertainty refinement module. Specifically, $\mathcal{L}_{bev}$ is the cross-entropy between the BEV semantic estimation $\mathbf{L}_{BEV}\in \mathbb{R} ^{N\times H \times W}$ from the satellite branch and the squeezed ground truth, while $\mathcal{L}_{co}$ is the cross-entropy between $\mathbf{L}_{coarse}$ and downsampled ground truth. These two losses are weighted by scale factor $\lambda_{bev}$ and $\lambda_{co}$, respectively. 
The total loss function can be expressed as follows:
\begin{equation}
\mathcal{L}=\mathcal{L}_{s c a l}^{g e o}+\mathcal{L}_{\text {scal }}^{s e m}+\mathcal{L}_{c e} + \lambda_{bev}\mathcal{L}_{bev} + \lambda_{co}\mathcal{L}_{co},
\end{equation}
where the scale factor ${L}_{bev}$ and $\lambda_{co}$ are set to 1 and 0.25. In addition, we employ the class weight refer to~\cite{voxformer}
.
\definecolor{color1}{RGB}{176, 36, 24}
\definecolor{color2}{RGB}{119,185,0}
\definecolor{color3}{RGB}{0, 0, 200}

\newcommand{\tbr}[1]{\textbf{\textcolor{color1}{#1}}}
\newcommand{\tbg}[1]{\textbf{\textcolor{color2}{#1}}}
\newcommand{\tbb}[1]{\textbf{\textcolor{color3}{#1}}}

\section{Experiments}
\begin{table*}[htb]
    \centering
    \newcommand{\clsname}[2]{
        \rotatebox{90}{
            \hspace{-6pt}
            \textcolor{#2}{$\blacksquare$}
            \hspace{-6pt}
            \renewcommand\arraystretch{0.6}
            \begin{tabular}{l}
                #1                                      \\
                \hspace{-4pt} ~\tiny(\semkitfreq{#2}\%) \\
            \end{tabular}
        }}
    \renewcommand{\tabcolsep}{2pt}
    \renewcommand\arraystretch{1.1}
    \scalebox{0.84}{
        \begin{tabular}{l|rr|rrrrrrrrrrrrrrrrrrrr}
            \toprule
            Method                               &
            \multicolumn{1}{c}{IoU}              &
            mIoU                                 &
            \clsname{road$\star$}{road}                 &
            \clsname{sidewalk$\star$}{sidewalk}         &
            \clsname{parking$\star$}{parking}           &
            \clsname{other-grnd.$\star$}{otherground}   &
            \clsname{building$\star$}{building}         &
            \clsname{car}{car}                   &
            \clsname{truck}{truck}               &
            \clsname{bicycle}{bicycle}           &
            \clsname{motorcycle}{motorcycle}     &
            \clsname{other-veh.}{othervehicle}   &
            \clsname{vegetation$\star$}{vegetation}     &
            \clsname{trunk}{trunk}               &
            \clsname{terrain$\star$}{terrain}           &
            \clsname{person}{person}             &
            \clsname{bicyclist}{bicyclist}       &
            \clsname{motorcyclist}{motorcyclist} &
            \clsname{fence}{fence}               &
            \clsname{pole}{pole}                 &
            \clsname{traf.-sign}{trafficsign}
            \\
            \midrule
            \multicolumn{3}{l}{\textit{LiDAR-based methods}}
            \\
            \hline
            LMSCNet \cite{lmscnet}   & 28.61          & 6.70           & 40.68          & 18.22          & 4.38           & 0.00           & 10.31          & 18.33          & 0.00           & 0.00          & 0.00          & 0.00           & 13.66          & 0.02           & 20.54          & 0.00          & 0.00          & 0.00          & 1.21          & 0.00          & 0.00          \\
            JS3C-Net \cite{js3cnet}  & 38.98          & 10.31          & 50.49          & 23.74          & 11.94          & 0.07           & 15.03          & 24.65          & 4.41           & 0.00          & 0.00          & 6.15           & 18.11          & 4.33           & 26.86          & 0.67          & 0.27          & 0.20 & 3.94          & 3.77          & 1.45          \\
            \hline
                        \multicolumn{3}{l}{\textit{Camera-based methods}}
            \\
            \hline

            MonoScene \cite{monoscene}   & 36.86          & 11.08          & 56.52          & 26.72          & 14.27          & 0.46           & 14.09          & 23.26          & 6.98           & 0.61          & 0.45          & 1.48           & 17.89          & 2.81           & 29.64          & 1.86          & 1.20          & 0.00          & 5.84          & 4.14          & 2.25          \\
            TPVFormer \cite{tpvformer}        & 35.61          & 11.36          & 56.50          & 25.87          & \tbg{20.60}        & 0.85           & 13.88          & 23.81          & 8.08           & 0.36          & 0.05          & 4.35           & 16.92          & 2.26           & 30.38          & 0.51          & 0.89          & 0.00          & 5.94          & 3.14          & 1.52          \\
            VoxFormer \cite{voxformer}          & \tbg{44.05} & 12.30          & 54.48          & 25.95          & 16.21          & 0.68           & \tbb{17.61}         & 25.82          & 5.48           & 0.61          & 0.50          & 3.86           & \tbb{24.22}         & \tbb{4.88}         & 29.24          & 1.88          & \tbr{3.13} & 0.00          & \tbb{7.88}          & \tbb{7.01}          & \tbb{4.20}          \\
            OccFormer \cite{occformer}          & 36.50          & \tbb{13.46}          & \tbb{58.85}          & 26.88          & \tbb{19.61}        & 0.31           & 14.40          & 25.09          & \tbr{25.53} & 0.81          & 1.19          & \tbb{8.52}           & 19.63          & 3.93           & \tbg{32.62} & {2.78}          & \tbg{2.82}          & 0.00          & 5.61          & 4.26          & 2.86          \\
            SurroundOcc \cite{surroundocc}          & 37.24          & 12.70          & \tbg{59.20} & \tbg{28.24} & \tbr{21.42} & \tbb{1.67}  & 14.94          & \tbb{26.26}          & 14.75         & \tbb{1.67}          & \tbb{2.37}          & 7.73           & 19.09          & 3.51           & \tbb{31.04}         & \tbb{3.60}          & \tbb{2.74}          & 0.00          & 6.65          & {4.53}          & 2.73          \\
            Symphonies \cite{symphonize}                            & \tbb{41.92}          & \tbg{14.89} & 56.35          & \tbb{27.58}          & 15.28          & \tbb{0.95}          & \tbg{21.64} & \tbg{28.68} & \tbg{20.44}          & \tbg{2.54} & \tbg{2.82} & \tbr{13.89} & \tbg{25.72} & \tbg{6.60} & 30.87           & \tbg{3.52} & 2.24          & 0.00          & \tbg{8.40} & \tbg{9.57} & \tbg{5.76} \\
            \hline
            \textbf{Ours}  & \tbr{45.01}       & \tbr{16.68} & \tbr{60.85}         & \tbr{31.53}         & 19.32        & \tbr{2.14}          & \tbr{26.05} & \tbr{32.17}& \tbb{20.30}         & \tbr{2.95} & \tbr{3.10} & \tbg{11.55} & \tbr{27.11} & \tbr{8.28} & \tbr{38.47}          & \tbr{3.66} &1.49          & 0.00          & \tbr{9.28} & \tbr{11.58} & \tbr{7.22} \\
            \bottomrule
        \end{tabular}
    }
    \caption{Quantitative results on SemanticKITTI \texttt{val} set. $\star$ denotes the scene layout structures.}
    \label{tab:sem_kitti_val}
\end{table*}

\begin{table*}[ht]
    \centering
    \newcommand{\clsname}[2]{
        \rotatebox{90}{
            \hspace{-6pt}
            \textcolor{#2}{$\blacksquare$}
            \hspace{-6pt}
            \renewcommand\arraystretch{0.6}
            \begin{tabular}{l}
                #1                                       \\
                \hspace{-4pt} ~\tiny(\sscbkitfreq{#2}\%) \\
            \end{tabular}
        }}
    \newcommand{\empa}[1]{\textbf{#1}}
    \newcommand{\empb}[1]{\underline{#1}}
    \renewcommand{\tabcolsep}{2pt}
    \renewcommand\arraystretch{1.2}
    \scalebox{0.84}
    {
        \begin{tabular}{l|rr|rrrrrrrrrrrrrrrrrr}
            \toprule
            \multicolumn{1}{c|}{Method}                                 &
            \multicolumn{1}{c}{IoU}                                     &
             \multicolumn{1}{c|}{mIoU}                                 &
            \multicolumn{1}{c}{\clsname{car}{car}}                      &
            \multicolumn{1}{c}{\clsname{bicycle}{bicycle}}              &
            \multicolumn{1}{c}{\clsname{motorcycle}{motorcycle}}        &
            \multicolumn{1}{c}{\clsname{truck}{truck}}                  &
            \multicolumn{1}{c}{\clsname{other-veh.}{othervehicle}}      &
            \multicolumn{1}{c}{\clsname{person}{person}}                &
            \multicolumn{1}{c}{\clsname{road$\star$}{road}}                    &
            \multicolumn{1}{c}{\clsname{parking$\star$}{parking}}              &
            \multicolumn{1}{c}{\clsname{sidewalk$\star$}{sidewalk}}            &
            \multicolumn{1}{c}{\clsname{other-grnd.$\star$}{otherground}}      &
            \multicolumn{1}{c}{\clsname{building$\star$}{building}}            &
            \multicolumn{1}{c}{\clsname{fence}{fence}}                  &
            \multicolumn{1}{c}{\clsname{vegetation$\star$}{vegetation}}        &
            \multicolumn{1}{c}{\clsname{terrain$\star$}{terrain}}              &
            \multicolumn{1}{c}{\clsname{pole}{pole}}                    &
            \multicolumn{1}{c}{\clsname{traf.-sign}{trafficsign}}       &
            \multicolumn{1}{c}{\clsname{other-struct.$\star$}{otherstructure}} &
            \multicolumn{1}{c}{\clsname{other-obj.}{otherobject}}
            \\
            \midrule
                        \multicolumn{3}{l}{\textit{LiDAR-based methods}}
            \\
            \hline

            SSCNet \cite{sscnet}                                    & 53.58  & 16.95        & 31.95 & 0.00        & 0.17        & 10.29        & 0.00         & 0.07        & 65.70 & 17.33 & 41.24 & 3.22        & 44.41 & 6.77       & 43.72 & 28.87 & 0.78         & 0.75        & 8.69         & 0.67         \\
            LMSCNet \cite{lmscnet}                                    & 47.35         & 13.65        & 20.91        & 0.00        & 0.00        & 0.26         & 0.58         & 0.00        & 62.95        & 13.51        & 33.51        & 0.20        & 43.67        & 0.33        & 40.01        & 26.80        & 0.00         & 0.00        & 3.63         & 0.00         \\
            \specialrule{0.7pt}{0pt}{0pt}
                        \hline
                        \multicolumn{3}{l}{\textit{Camera-based methods}}
            \\
            \hline

            MonoScene \cite{monoscene}                                 & 37.87           & 12.31        & 19.34        & 0.43        & 0.58        & 8.02         & 2.03         & 0.86        & 48.35        & 11.38        & 28.13        & 3.32        & 32.89        & 3.53        & 26.15        & 16.75        & 6.92         & 5.67        & 4.20         & 3.09         \\
            TPVFormer \cite{tpvformer}                                & \tbb{40.22}      & 13.64        & 21.56        & \tbb{1.09}        & 1.37        & 8.06         & 2.57         & 2.38        & 52.99        & 11.99        & \tbb{31.07}        & 3.78        & \tbb{34.83}        & 4.80        & 30.08        & 17.52        & 7.46         & 5.86        & 5.48         & 2.70         \\
            VoxFormer \cite{voxformer}                                  & 38.76         & 11.91        & 17.84        & \tbg{1.16}        & 0.89        & 4.56         & 2.06         & 1.63        & 47.01        & 9.67         & 27.21        & 2.89        & 31.18        & 4.97        & 28.99        & 14.69        & 6.51         & 6.92        & 3.79         & 2.43         \\
            OccFormer \cite{occformer}                                & 40.27        & \tbb{13.81}        & \tbb{22.58}        & 0.66        & 0.26        & 9.89         & 3.82         & 2.77        & 54.30        & \tbg{13.44}        & 31.53        & 3.55        & 36.42 & 4.80        & \tbb{31.00}        & \tbg{19.51} & \tbb{7.77}         & \tbb{8.51}        & 6.95         & 4.60         \\
            Symphonies \cite{symphonize}                             & \tbg{43.41}  & \tbg{17.82} & \tbg{26.86} & \tbr{4.21} & \tbr{4.90}& \tbg{14.20} & \tbr{7.76} & \tbr{6.57} & 57.30& \tbr{13.58} & \tbg{35.24} & \tbb{4.57} & \tbg{39.20}       &\tbr{7.95} &\tbg{34.23} & \tbb{19.19}        & \tbg{14.04} & \tbg{16.78} & 8.23& \tbr{6.04} \\
            GaussianFormer \cite{gaussianformer}&35.38&12.92&18.93&1.02&\tbg{4.62}&\tbr{18.07}&\tbg{7.59}&\tbb{3.35}&45.47&10.89&25.03&\tbr{5.32}&28.44&\tbb{5.68}&29.54&8.62&2.99&3.32&\tbr{9.51}&\tbb{5.14}
            \\
            \hline
\textbf{Ours}&\tbr{46.35}&\tbr{18.30}&\tbr{27.80}&0.91&\tbb{2.55}&\tbb{10.73}&\tbb{5.67}&\tbg{4.28}&\tbr{61.04}&\tbb{13.21}&\tbr{37.00}&\tbg{5.07}&\tbr{43.05}&\tbg{7.46}&\tbr{38.98}&\tbr{24.87}&\tbr{15.75}&\tbr{16.90}&\tbg{8.85}&\tbg{5.33}
            \\
            \bottomrule
        \end{tabular}
    }
    \caption{Quantitative results on SSCBench-KITTI-360 \texttt{test} set. $\star$ denotes the scene layout structures.}
    \label{tab:kitti_360_test}
    \vspace{-3mm}
\end{table*}
To fairly evaluate the performance of SGFormer, we conduct experiments on the SemanticKITTI~\cite{semantickitti} and SSCBench-KITTI-360~\cite{sscbench} datasets. In Sec~\ref{sec:main_result}, we compare our method against existing approaches on both datasets. We mark the top-3 results of camera-based methods in \tbr{red}, \tbg{green}, and \tbb{blue}. In Sec~\ref{sec:ablation}, we present ablation studies to demonstrate the effectiveness of each module and selection. Finally, in Sec~\ref{sec:visulization}, we show our visualization results.
\subsection{Datasets}
The SemanticKITTI dataset contains 22 sequences. Among them, 11 sequences are used for training, 1 sequence is for validation, and 10 sequences are for testing. It provides ground images with the shape of $1226 \times 370$. Meanwhile, corresponding satellite image data are provided by~\cite{beyond}. Each satellite image has the size of $512 \times 512$, with a scale factor of 0.2 meters per pixel. In this paper, due to the lack of GPS information, we only use 10 sequences to train the model (without sequence 03) and evaluate the performance on the validation set.

SSCBench-KITTI-360 dataset contains 9 sequences, 7 sequences are used for training, 1 sequence for validation, and 1 for test. The input size of the ground image is $1480 \times 376$. Since the dataset does not include satellite data, we obtained corresponding satellite images from Google Maps~\cite{google_maps_api} using the ground-truth poses provided. The satellite image settings are consistent with those used in SemanticKITTI.

Both SemanticKITTI and SSCBench-KITTI-360 datasets provide the voxelized point clouds with labels as ground truth, measuring the 3D volume with a range of $51.2m \times 51.2m \times 6.4m$. The dimension of voxel grids is $256 \times 256 \times 32$, making each grid cell a 0.2-meter cube. In our evaluation, we report the intersection over union (IoU) for geometry performance and mean IoU (mIoU) metrics for semantic performance, aligning with mainstream works.

\subsection{Implementation Details}
We train our SGFormer for 25 epochs on 3 NVIDIA 3090 GPUs, with a batch size of 3. We employ AdamW optimizer~\cite{adamw} with an initial learning rate of 4e-4, with a weight decay of 0.01. We employ the cosine learning rate strategy to reduce the learning rate during the training. Furthermore, the feature dimension is set to 128, and each batch consumes around 20GB of GPU memory. 
\begin{table*}[t]
\centering
\scalebox{0.9}{
\begin{tabular}{ccc|cccc|cc}
\toprule
Sat.-branch & Sat.-corr. & Fusion & IoU     & mIoU    & Global   & Detail &Params (M) &Memory (M)  \\ \midrule
\ding{55}           & \ding{55}          & \ding{55}   & 44.53 & 14.80 & 26.13 & 8.21 &93.49 & 15424\\
\ding{51}         &   \ding{55}     &    \ding{55}       & 44.00& 15.01 & 26.54 & 8.26&126.43 &18263\\
\ding{51}          &\ding{51}       &     \ding{55}        & 43.90& 15.59 & 28.43 & 8.10 &126.53 &18967 \\
\ding{51}         &   \ding{55}      & \ding{51}        & 44.74 & 15.81& 27.50& 9.01 &126.89 &19123\\
\ding{51}          & \ding{51}        & \ding{51}        & \bf{45.01} & \bf{16.68} & \bf{29.31}& \bf{9.29} &126.99 & 19865\\
\bottomrule
\end{tabular}}
\vspace{-1mm}
\caption{Ablation on main components of SGFormer.}
\label{tab:ablation_1}
\vspace{-5mm}
\end{table*}

\begin{table}[t]
\centering
\scalebox{0.85}{
\begin{tabular}{ccccc}
\toprule
\multicolumn{1}{c|}{}           & IoU     & mIoU    & Global   & Detail  \\ \midrule
\multicolumn{5}{c}{\bf{w/o Sat.-corr.}}                                       \\ \midrule
\multicolumn{1}{c|}{w/o  noise} & 44.74 & 15.81& 27.50 & 9.01\\
\multicolumn{1}{c|}{±5m}        & 44.83 & 15.10 & 26.08 & 8.70 \\ \midrule
\multicolumn{5}{c}{\bf{with Sat.-corr.}}                                         \\ \midrule
\multicolumn{1}{c|}{w/o  noise} & 45.01 & 16.68& 29.31 & 9.29 \\
\multicolumn{1}{c|}{±5m}        & 45.24 &  15.96&   28.75&    8.53\\ 
\bottomrule
\end{tabular}}
\caption{Ablation on localization noise.}
\label{tab:ablation_2}
\vspace{-4mm}
\end{table}

\subsection{Main Results}
\label{sec:main_result}
We compare our method with state-of-the-art on both semanticKITTI and SSCBench-KITTI-360 datasets. The baseline methods include camera-based methods, as well as LiDAR-based methods. Since SSC performance is sensitive to depth quality, we re-evaluated the depth-based method Voxformer~\cite{voxformer} and Symphonize~\cite{symphonize} using our generated depth data (the depth estimation method is the same). The results for other methods are obtained from~\cite{symphonize,sscbench}. Table \ref{tab:sem_kitti_val} and Table \ref{tab:kitti_360_test} show the quantitative results. It is demonstrated that our method achieves superior performance on both geometry (IoU) and semantics predictions (mIoU). Regarding semantic prediction, our method achieved 16.68 mIoU on the SemanticKITTI dataset and 18.30 mIoU on SSCBench-KITTI-360, significantly outperforming all other methods. For occupancy prediction, our method achieved 45.01 IoU on the SemanticKITTI dataset, surpassing all previous approaches, including VoxFormer, which uses additional steps to estimate occupancy, and all LiDAR-based methods. On the SSCBench-KITTI-360 dataset, our method achieved 46.55 IoU, outperforming all other camera-based methods and achieving results that are also quite comparable to LiDAR-based methods.
For a detailed analysis of specific labels, our method performs significantly better than other camera-based approaches in predicting scene layouts. On the SSCBench-KITTI-360 dataset, SGFormer is the only camera-based method to achieve performance comparable to LiDAR-based methods in scene layout estimation. This improvement is due to the additional satellite input, which provides a more complete view of these objects through aerial observations. Moreover, when evaluating the performance of small or dynamic objects, we find that our method also achieves excellent performance. The adaptive fusion module effectively mitigates the negative effects of satellite observations on those small objects. A more detailed analysis of this part is provided in the ablation study section~\ref{sec:ablation}.

\subsection{Ablation Study}
\label{sec:ablation}
In this section, we conduct the ablations on the SemanticKITTI dataset, mainly analyzing the effectiveness of our core designs and the impact of localization noise. In addition to the IoU and mIoU, we also report the mean prediction accuracy (mIoU) of different semantic categories for further analysis. Specifically, we set those scene layout structures as the "global" category, while other dynamic, small objects are regarded as the "detail" category, reporting the performance of detail estimation. 
\begin{figure*}
  \centering
  \includegraphics[width=\linewidth]{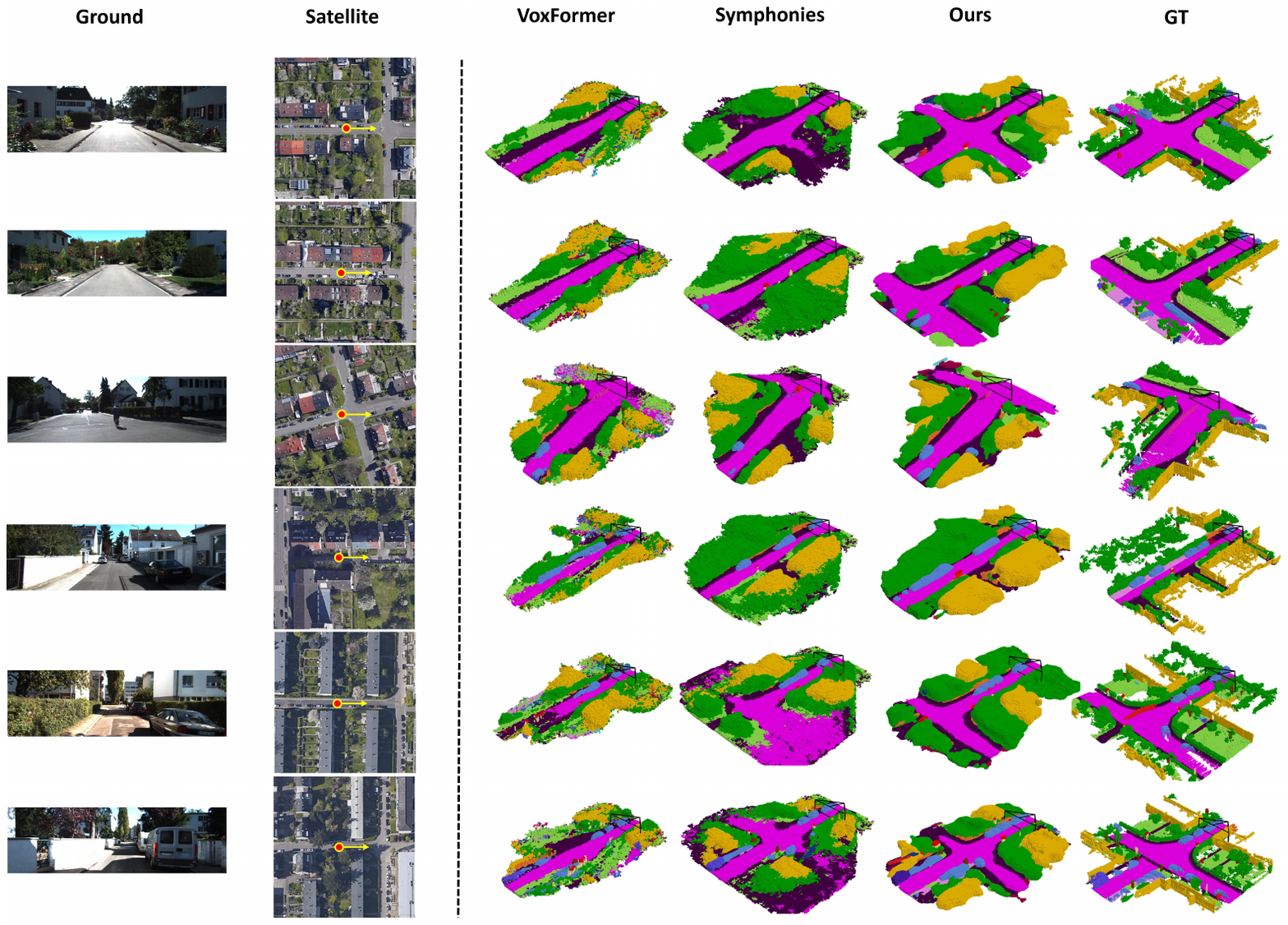}
  \vspace{-11mm}
  \caption{\textbf{Scene Semantic Completion on SemanticKITTI~\cite{semantickitti}.} \textbf{Left}: we show satellite-ground image pairs and indicate the vehicle's travel direction as a yellow arrow. \textbf{Right}: we qualitatively compare scene semantic completion results from SGFormer and other baselines, where SGFormer can produce more complete and accurate semantic reconstruction relying on the satellite-ground fusion.} 
  \label{fig:visualization}
  \vspace{-5mm}
\end{figure*}

\subsubsection{Effectiveness on Core Designs}
Table \ref{tab:ablation_1} presents the ablation on three core designs: the satellite branch (sat.-branch. in the table), satellite correction strategy (sat.-corr. in the table), and adaptive fusion module (fusion in the table). A single ground-branch approach is set as the baseline. 

\noindent \textbf{Satellite branch.} As shown, adding a new branch with satellite inputs can improve the overall performance. However, the improvement is slight in both categories, and it leads to a degradation in IoU performance. These observations indicate two key points. First, adding satellite observations does enhance the prediction of scene distribution, but misalignment issues such as satellite image localization errors and vertical occlusions can limit this improvement. Second, the lack of observations on dynamic objects, small objects, and occupancy information in satellite images leads to negative effects on predictions for these “detailed" objects, as well as the geometry prediction. 

\noindent \textbf{Satellite correction strategy.} As seen in the table, employing our satellite correction strategy can significantly improve the prediction accuracy of scene layout structures in the "global" category, increasing from 26.54 to 28.43. This improvement demonstrates that our satellite correction strategy can address the misalignment issues. The ground-view observation assists the satellite branch in completing the scene layout.

\noindent \textbf{Adaptive fusion module.} The ablation also indicates that using our adaptive fusion module can significantly enhance the accuracy for small and dynamic objects. As shown in the table, with the fusion module, the mIoU of "detail" category is increased from 8.26 to 9.01. Furthermore, adding the fusion module will also lead to an improvement of IoU. 

It is noted that combining all these components yields a substantial performance improvement, as shown in the last row of the table.

\subsubsection{Impact of Localization Noise}
We conduct the experiment to explore the impact of the localization noise, as shown in table \ref{tab:ablation_2}. We add a random noise of $\pm5$ meters along both latitude and longitude to align the possible localization errors in the real world. Overall, our method is sensitive to localization noise: adding a 5-meter noise results in a decrease of 4.3\% and 4.5\% mIoU for SGFormer with and without the satellite correction strategy, respectively.  Our satellite correction does indeed reduce the negative impact caused by localization noise, which is particularly evident in the scene layout performance. Without satellite correction, the mIoU for the "global" category dropped to 26.08, even lower than the baseline value in table \ref{tab:ablation_1}, indicating that satellite observations almost no longer provide a positive effect. However, when using our correction strategy, the performance decline in scene layout prediction is significantly mitigated.

\subsection{Qualitative Comparisons}
\label{sec:visulization}
Figure~\ref{fig:visualization} shows the qualitative comparisons between our SGFormer and two baseline methods, VoxFormer~\cite{voxformer} and Symphonies\cite{symphonize} on SemanticKITTI. As shown, the results predicted by our method are significantly better than the other two methods. Overall, our output shows a more accurate scene layout, particularly in complex scenes such as cross-road. The other two methods struggle to reconstruct these complex scenes completely, whereas our method handles this situation effectively.
Considering the detail part, VoxFormer produces the scene with radial artifacts, while Symphonies performs somewhat better in this regard but still has many cluttered areas due to occlusions. Our reconstructed results not only avoid these issues but also produce clearer boundaries between different semantics. These comparisons demonstrate the superiority of our method both global and local scale. 

\section{Conclusion}
In this paper, we introduce SGFormer, the first satellite-ground cooperative approach with a tightly coupled dual-branch framework design for urban semantic scene completion. SGFormer effectively fuses observations from ground-view and satellite-view images, enabling joint consideration of global scene layout and local details for occupancy grid prediction. Experimental results show that our method surpasses state-of-the-art camera-based approaches and achieves performance on par with LiDAR-based methods across many metrics. These findings suggest that integrating satellite images into the SSC task offers a cost-effective yet highly promising solution. We hope our work inspires further research in this community.



\section*{Acknowledgment}
This work was partially supported by NSF of China (No. 62425209). The author Junjie Hu acknowledges support from the Guangdong Natural Science Fund under Grant 2024A1515010252. The authors thank Shang Liu for providing the figures used in this paper. Special thanks also go to Bo Wen, Xingyuan Yu, Han Zhang, and Ran Zuo for their invaluable support.
{
    \small
    \bibliographystyle{ieeenat_fullname}
    \bibliography{main}
}


\end{document}